\documentclass[letterpaper]{article} 
\usepackage[]{aaai24}  
\usepackage{times}  
\usepackage{helvet}  
\usepackage{courier}  
\usepackage[hyphens]{url}  
\usepackage{graphicx} 
\usepackage{natbib}  
\usepackage{caption} 
\usepackage{algorithm}
\usepackage{algorithmic}
\usepackage{booktabs}       
\usepackage{amsfonts}       
\usepackage{nicefrac}       
\usepackage{microtype}      
\usepackage{xcolor}         
\usepackage{enumitem}
\usepackage{makecell}
\usepackage{amssymb}
\usepackage{amsthm}
\usepackage[switch]{lineno}
\usepackage{amsmath} 
\usepackage{subfigure}
\usepackage{multirow}

\usepackage{colortbl}

\definecolor{gray}{RGB}{222,222,222}
\allowdisplaybreaks

\usepackage{bbding} 
\usepackage{url}

\usepackage{newfloat}
\usepackage{listings}
\DeclareCaptionStyle{ruled}{labelfont=normalfont,labelsep=colon,strut=off} 
\floatstyle{ruled}
\newfloat{listing}{tb}{lst}{}
\floatname{listing}{Listing}
\title{Collaborative Chinese Text Recognition with Personalized Federated Learning}
\author{
    Shangchao Su,
    Haiyang Yu,
    Bin Li \thanks{Corresponding author},
    Xiangyang Xue
}
\affiliations{
    Fudan University, Shanghai, China\\
    \{scsu20, hyyu20, libin, xyxue\}@fudan.edu.cn
}

\usepackage{bibentry}

\begin{document}
\maketitle

\begin{abstract}
In Chinese text recognition, to compensate for the insufficient local data and improve the performance of local few-shot character recognition, it is often necessary for one organization to collect a large amount of data from similar organizations. However, due to the natural presence of private information in text data, such as addresses and phone numbers, different organizations are unwilling to share private data. Therefore, it becomes increasingly important to design a privacy-preserving collaborative training framework for the Chinese text recognition task. In this paper, we introduce personalized federated learning (pFL) into the Chinese text recognition task and propose the pFedCR algorithm, which significantly improves the model performance of each client (organization) without sharing private data. Specifically, pFedCR comprises two stages: multiple rounds of global model training stage and the the local personalization stage. During stage 1, an attention mechanism is incorporated into the CRNN model to adapt to various client data distributions. Leveraging inherent character data characteristics, a balanced dataset is created on the server to mitigate character imbalance. In the personalization phase, the global model is fine-tuned for one epoch to create a local model. Parameter averaging between local and global models combines personalized and global feature extraction capabilities. Finally, we fine-tune only the attention layers to enhance its focus on local personalized features. The experimental results on three real-world industrial scenario datasets show that the pFedCR algorithm can improve the performance of local personalized models by about 20\% while also improving their generalization performance on other client data domains. Compared to other state-of-the-art personalized federated learning methods, pFedCR improves performance by 6\% $\sim$ 8\%.
\end{abstract}

    \section{Introduction}

In the era of deep learning, OCR (Optical Character Recognition) technology is becoming increasingly popular and is widely used in scenarios such as document scanning~\cite{narang2019devanagari,hsu2020intelligent}  and invoice recognition~\cite{blanchard2019automatic,ming2003research}. In these scenarios, Latin text recognition has become more and more mature due to its simple characters and long research history. However, Chinese text recognition still requires further exploration due to factors such as complex characters and numerous categories~\cite{yu2021benchmarking}.

\begin{figure}[t]
 \centering
 \includegraphics[width=0.9\linewidth]{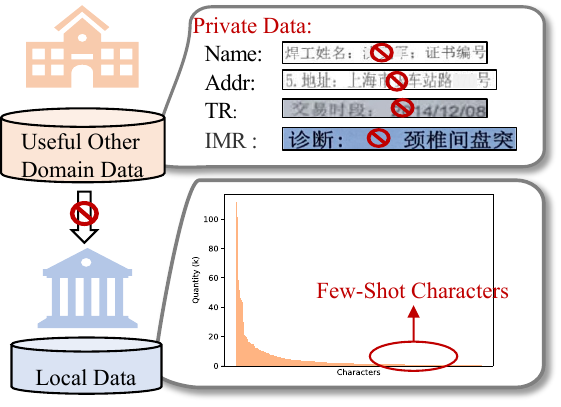}
 \centering
 \caption{
A dilemma in Chinese text recognition that we focus on: 1) There are a large number of few-shot characters in Chinese text data, making it difficult to improve the performance of few-shot character recognition in a single organization without data sharing. 2) Due to privacy concerns, different organizations are unwilling to share their text data.
 }
 \label{issue}
 \end{figure}

In the existing Chinese text recognition scenario, one organization often needs to collect a large amount of related data from other organizations in order to compensate for the insufficient local data during training. For example, if a hospital is conducting medical record recognition, collecting medical records from other hospitals may help improve the performance of the local model. Nevertheless, since medical record data contains strong privacy information, third-party organizations are often unwilling to share their private data. As shown in Figure~\ref{issue}, personal information such as name, address, transaction records (TR), and individual medical records (IMR) cannot leave the local environment, otherwise it would violate regulatory policies.

Furthermore, unlike Latin text recognition, Chinese text recognition involves numerous character categories~\cite{yu2021benchmarking}. However, commonly used characters in a single scenario are relatively few, resulting in a significant long-tail distribution of character frequency, as illustrated in Figure~\ref{issue}. A small number of characters account for a large amount of data. This makes the performance of a locally trained model on few-shot characters poor. In addition, data from different scenarios often contain different high-frequency character types, making data from different parties complementary.

Given the aforementioned practical demands, it is becoming increasingly important to develop a privacy-preserving collaborative training framework for the Chinese text recognition task. We note that the recently popular federated learning techniques hold promise for addressing these demands. Federated learning~\cite{mcmahan2017communication} is a collaborative learning method that can protect privacy and allow clients to jointly build models without sharing raw data. In the standard federated learning setting~\cite{mcmahan2017communication,conf/mlsys/LiSZSTS20,karimireddy2020scaffold,lin2020ensemble}, the server is responsible for aggregating client models and sharing the aggregated model (global model) in each communication round. The clients receive the global model and conduct local training before uploading the trained models. The global model gradually converges after multiple rounds of communication. As the differences in client data distribution increase, a single global model cannot fit all data distributions. Therefore, emerging personalized federated learning (pFL)~\cite{li2021fedbn,collins2021exploiting,huang2021personalized,fallah2020personalized} techniques allow different clients to obtain different models, thus mitigating the impact of data heterogeneity.
 
In this paper, we aim to explore collaborative Chinese text recognition. We have two main goals: 1) to enable multiple clients, such as companies with invoice recognition needs, to collaborate on model training without sharing sensitive text fragments on the client. 2) to enhance the performance on few-shot characters, compensating for the inadequacy of traditional single-party training.

Specifically, we introduce personalized federated learning into the task of Chinese text recognition and point out two challenges encountered when applying pFL to Chinese text recognition. And then we propose the pFedCR algorithm.  Our framework primarily consists of two stages: the global model training (GMT) stage and the local personalization (LPS) stage. In the GMT stage, the goal is to acquire an initial model capable of handling various types of characters and possessing adaptive feature extraction capabilities for different distributions through multiple rounds of collaborative training. To achieve this, we introduce an attention mechanism with parameters $e$ to the CRNN model $\omega$. This empowers the attention layer to adapt to the data distribution of different clients during the global model's learning process. Additionally, leveraging the inherent characteristics of character data, which can be easily used to generate a large number of virtual data composed of random characters, we create a balanced character dataset on the server. After each round of parameter averaging, we fine-tune the global model using this balanced dataset. During the client training phase, to preserve the balance of the classification head, we freeze the last layer of the model.

In the LPS stage, the goal is to personalize the global model from stage 1 to better fit local features while minimizing disruption to the initial global model. To achieve this objective, we conduct another epoch of fine-tuning the global model to obtain a local model. Then, we perform parameter averaging between the local and global models to obtain $\{\omega_{k,p}, e_{k,p}\}$, where $\omega_{k,p}$ has gained to some extent both personalized feature extraction capabilities and global model feature extraction capabilities. Finally, we freeze $\omega_{k,p}$ and fine-tune $e_{k,p}$ with regularization, allowing the attention layers to emphasize local personalized features.

We conduct extensive experiments on public datasets and two real-world scenarios datasets. The results demonstrate that introducing pFL can significantly improve the local model performance of clients by 20\%. Moreover, the proposed pFedCR outperforms all competitors with a performance improvement of nearly 8\%. Additionally, we observe that the personalized model obtained by one client through collaborative learning has significantly improved generalization performance on other clients. pFedCR demonstrates the powerful potential of pFL in Chinese text recognition scenarios.  Also, we demonstrate that the standard parameter-sharing mechanism exhibits a higher level of privacy preservation in character recognition compared to image recognition. Extracting semantic information such as telephone numbers from model parameters is significantly more challenging in text data than deducing specific class images from image classification models. 

Our contributions are summarized as follows: 
1) We introduce pFL into the Chinese text recognition task for the first time to address the sharing-vs-privacy dilemma. 
2) We propose the pFedCR algorithm to address the character imbalance and non-iid problems faced by pFL in Chinese text recognition.
3) We thoroughly validate the effectiveness of pFedCR on both public datasets and self-collected real-world datasets, demonstrating its advantages over local training and existing pFL approaches.

\section{Related Work}
\subsection{Optical Character Recognition}
Optical Character Recognition (OCR) has been extensively researched for decades due to its wide applications. However, most of the previous text recognition methods focus on Latin text recognition~\cite{wang2021two,yan2021primitive,li2021trocr}. CRNN~\cite{shi2016end} was one of the earlier methods that proposed to combine CNN and RNN to extract visual features from text images and employed the CTC loss~\cite{graves2006connectionist} to optimize the model. Recently, some methods have introduced semantic models to further enhance the performance of text recognition (\textit{e.g.}, SEED~\cite{qiao2020seed}, ABINet~\cite{fang2021read}, and SRN~\cite{yu2020towards}. However, few methods have been proposed specifically for Chinese text recognition though there exist several challenges to be resolved, such as a large number of characters and vertical texts.



\subsection{Personalized Federated Learning}  The purpose of Federated Learning~\cite{mcmahan2017communication}  is to enable multiple clients to jointly model without sharing raw private data. Personalized federated learning (pFL) is proposed to address the non-iid issue, which allows different clients to obtain personalized models instead of a unified global model. Motivated by multi-task learning~\cite{caruana1997multitask}, \cite{DBLP:conf/nips/SmithCST17} proposes MOCHA using a primal-dual optimization method. pFedAvg \cite{fallah2020personalized} uses a meta-learning approach to learn a good initial global model that performs well on each client. pFedMe~\cite{DBLP:conf/nips/DinhTN20} and Ditto~\cite{li2021ditto} decouple personalized model optimization from global model learning by adding regularization. FedAMP \cite{huang2021personalized} proposes attentive message passing to facilitate similar clients to collaborate more. FedFomo~\cite{DBLP:conf/iclr/ZhangSFYA21} calculates the optimal weighted combination of local models as a personalized federated update. FedALA~\cite{zhang2022fedala} adaptively selects global model parameters for local model initialization through local training. The above works use a single network to learn general and personalized knowledge at the same time. There are also some works that try to separate personalized parameters from the network for local knowledge learning, and only share the non-personalized part of the models to the server during federated learning. \cite{DBLP:journals/corr/abs-1912-00818,DBLP:conf/icml/CollinsHMS21} proposes a base + personalization layer approach, the personalization layer always stays locally on the client. LG-FedAvg \cite{liang2020think} uses lower layers as local encoders which are subsequently connected to a global classifier.

\subsection{Text Recognition with FL}
In some Latin text recognition tasks, previous works~\cite{zhang2020fedocr,ren2020privacy} have attempted to address text recognition using standard federated learning. In the Chinese context, there is only one work~\cite{zhu2019federated} that attempts to validate the effectiveness of the standard federated learning framework in Chinese text recognition. These early attempts also reveal the demand for federated learning in text recognition tasks.

\section{Method}
In this section, we first introduce the problem setting and analyze the challenges of the problem, then present our proposed federated Chinese text recognition framework, pFedCR.

\subsection{Problem Setting}


Suppose $\mathcal{D}_k$ is the dataset of the $k$-th client (participant), which may be a company with a demand for text recognition. In standard federated learning, the objective is to minimize:
\begin{linenomath}
\begin{align}
F(\omega)=\frac{1}{K} \sum_{k \in S} F_k(\omega)
\end{align}
\end{linenomath}
where $S$ is the client set, $F_k(\omega)$ is the local loss function:
\begin{linenomath}
\begin{align}
F_k(\omega)=\frac{1}{\left|\mathcal{D}_k\right|} \sum_{\left(x_j, y_j\right) \in \mathcal{D}_k} f\left(\omega, x_j, y_j\right)
\end{align}
\end{linenomath}

However, due to the distribution differences among different clients, a single global model is difficult to adapt to all distributions. Therefore, the goal of personalized federated learning is to improve the performance of each client in collaborative learning:
\begin{linenomath}
\begin{align}
\nonumber
F(\omega,\{\omega_k\}_{k=1}^K)=\frac{1}{K} \sum_{k \in S} F_k(\omega_k)+\mathcal{R}(\omega,\omega_1,\cdots,\omega_K)
\end{align}
\end{linenomath}
where $\omega_k$ is the $k$th local model. The purpose of the regularization term is to limit the distance between the global model and the local models, then the local models can learn part of the general knowledge while learning personalized knowledge.  

In practical applications, CRNN~\cite{shi2016end} has become the most popular Chinese text recognition model due to its balance between performance and speed, which combines convolutional layers to extract features, recurrent layers to capture temporal dependencies, and a final fully connected layer for classification. Therefore, in this paper, we chose CRNN as the model architecture for all clients. 

\subsection{Challenges}
\begin{figure}[t]
 \centering
 \includegraphics[width=1.\linewidth]{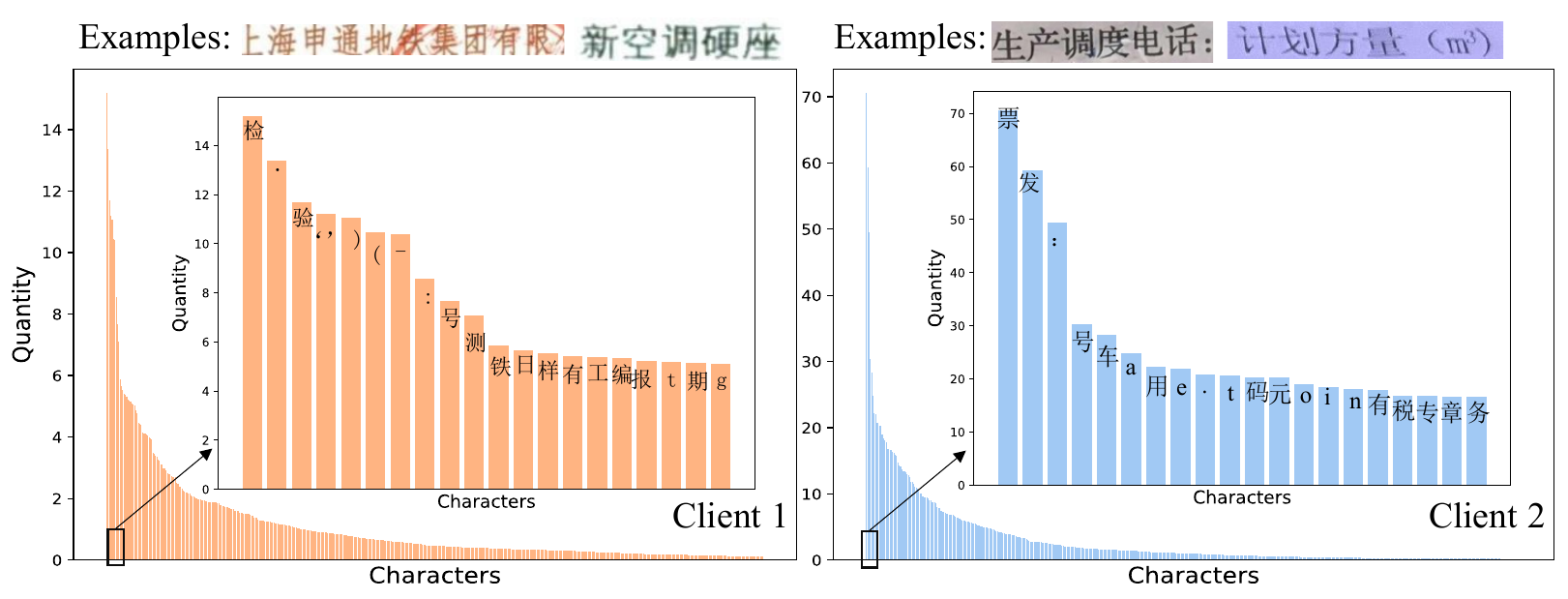}
 \caption{
The character imbalance and non-iid problems among different clients, the unit of quantity is k.
 }
 \label{dataset}
 \end{figure}

We summarize the challenges faced when implementing pFL for Chinese text recognition.
\begin{itemize}[noitemsep, leftmargin=*]
 \item\textbf{1) Imbalance.} The first one is the character imbalance problem.  As shown in Figure~\ref{dataset}, we present the statistical information of two clients' data, where `quantity' represents the number of images for each character. It can be observed that the text recognition dataset suffers from a severe character imbalance issue, exhibiting a long-tail distribution across the entire dataset.
\item\textbf{2) Non-iid.} The other is the non-iid problem. It is evident that the high-frequency character categories differ significantly among clients, particularly for Chinese characters due to diverse business scenarios. Furthermore, the data style differences caused by image backgrounds and fonts exacerbate the difficulty of conducting federated learning in this setting.
\end{itemize}

  \begin{figure}[t]
 \centering
 \includegraphics[width=1.\linewidth]{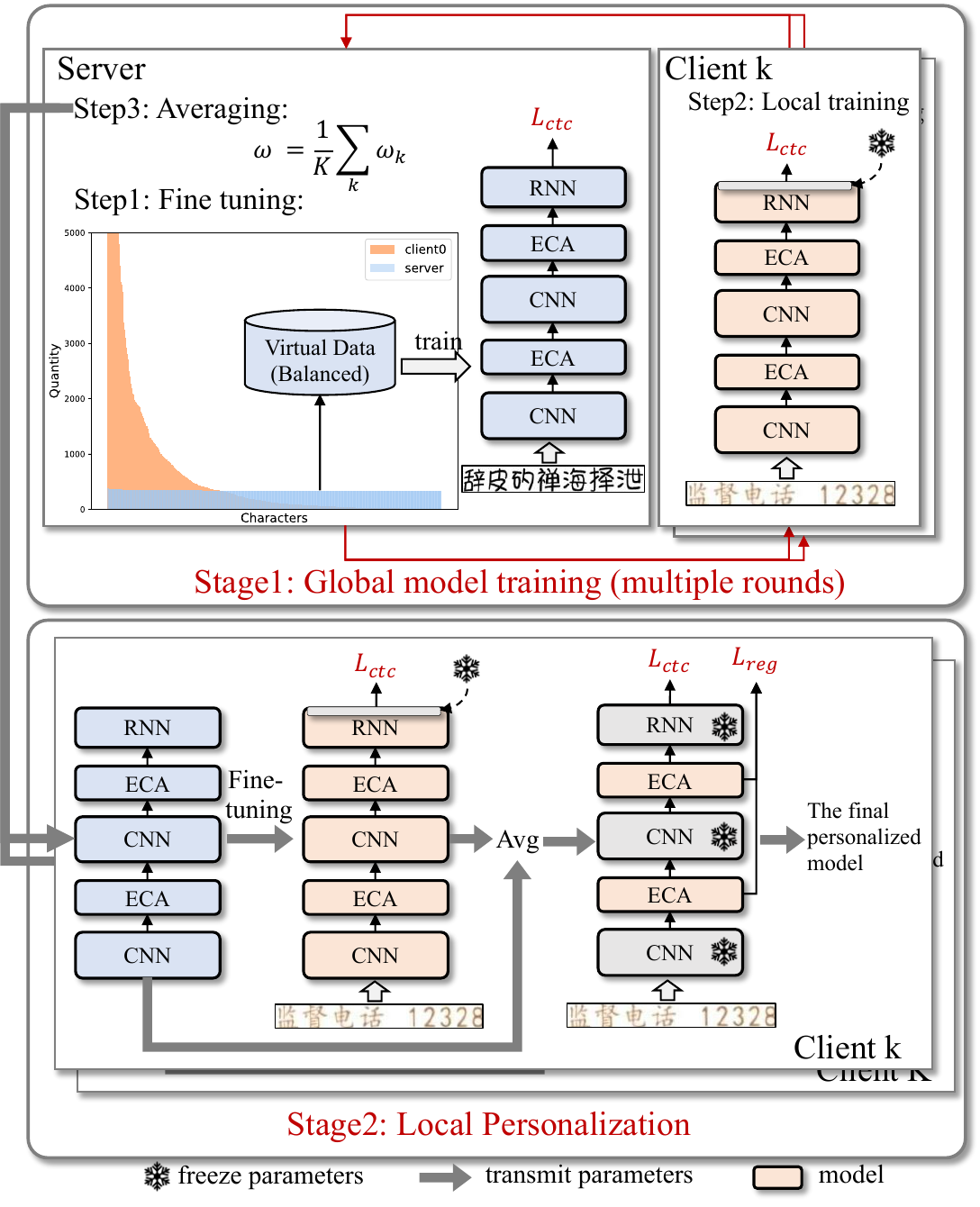}
 \caption{
An overview of pFedCR, which consists of two stages: global model training and personalization.
 }
 \label{framework}
 \end{figure}

 \subsection{pFedCR}
 

In general, our approach consists of two stages: the global model training stage and the personalization stage. Below, we will detail each of these two components.

\textbf{Global model training.}
Due to the non-iid problem in the client-side dataset, which can significantly disrupt feature extraction, we need different clients to focus more on local personalized features based on the global model. Therefore, we incorporate an attention mechanism into the global model, enabling it to adapt to local features while retaining as much knowledge from the global model as possible. The attention mechanism will play a crucial role in the personalized stage; however, for now, we are treating it as an integral part of the global model, participating in the federated learning process.

 Let the parameters of the CRNN be represented as $\omega$, and the parameters of the attention mechanism as $e$. We define the objective for this stage:
\begin{linenomath}
 \begin{align}
\label{pflpfl}
 \omega^*,e^* = \underset{\omega,e}{\arg \min } \frac{1}{K} \sum_{k \in S} F_k(\omega,e)
\end{align}
\end{linenomath}

where, $\{\omega^*,e^*\}$ is the global model. For the attention mechanism, we employ Efficient Channel Attention (ECA)~\cite{wang2020eca} layers. The ECA layer is defined as $\mathcal{A}(\cdot)$ which considers the interactions between each channel and its $r$-nearest neighbor channels. Given an input feature $Z\in\mathbb{R}^{B\times C \times H \times W}$, the mechanism of ECA is:
\begin{linenomath}
\begin{align}
\mathcal{A}(Z) & =Z\sigma(\operatorname{Conv1D}(\operatorname{GAP}(Z))) 
\end{align}
\end{linenomath}
where $\operatorname{Conv1D}$ denotes the $\operatorname{1D}$  convolution with a
kernel of shape $r$ on the channel dimension, $\operatorname{GAP}$  denotes the global average pooling operation. The hyperparameter $r$ is adaptively set to $r=\left|(\log _2(C)+1)/2\right|_{o d d}$, where $|x|_{o d d}$ denotes the odd number closest to $x$. In the experimental section, we will demonstrate that the improvement brought by the attention layer to federated learning is significantly higher than the improvement it brings to the local model trained within one organization. Furthermore, it is worth noting that the increase in the number of model parameters after adding the attention layer can be very small. Taking CRNN as an example, the proportion of attention layer parameters to the total number of model parameters is 1.3e-5.

One direct approach to solve Eq~\ref{pflpfl} is to use the FedAvg algorithm. However, in the context of this paper, significant variations in character distributions among different clients result in a pronounced character imbalance issue for the global model. In image classification tasks, the inability to access data for all categories on the server makes it challenging to address this problem during aggregation. Fortunately, for character data, which relies on a fixed dictionary, we can generate arbitrary controllable virtual samples on the server side using public tools, like Text Render \footnote{https://github.com/Sanster/text\_renderer}.

Note that when generating virtual data, there is no need to preserve the semantic information of the text, allowing for arbitrary combinations. Therefore, we predefine the quantity of each character and randomly combine them, maintaining a balance between the number of different types of characters, as shown in Figure~\ref{framework}. Then, on the server, we first fine-tune the global model using the virtual data (Step1 in Figure~\ref{framework}). 

When the global model is sent to the client for local training (Step2), to alleviate the impact of class imbalance during local training, we freeze the parameters of the last embedding layer in the local model. The loss function for local training is $L_{ctc}$, which is the standard CTC Loss~\cite{graves2006connectionist} for Chinese text recognition. Finally,  the server averages the client models' parameters: $\omega = \frac{1}{K}\sum_k\omega_k, e = \frac{1}{K}\sum_ke_k$. 

It is noteworthy that in this stage, there is an exchange of model parameters between the server and the clients. In traditional image classification tasks, this could potentially pose a privacy risk, as the model parameters themselves might be reverse-engineered to deduce class-specific samples~\cite{yin2020dreaming}. However, the privacy concern in this paper pertains to specific semantic information contained within the text, such as telephone numbers, which is distinct from simple class information. Reversely inferring this type of information through model parameters is considerably challenging. We will validate this aspect in the experimental section.

\textbf{Local Personalization.} 
After completing Stage 1, we obtain a global model $\{\omega^*, e^*\}$. In Stage 2, we proceed with further personalization. Firstly, to enhance the extraction ability of local personalized information, we perform a fine-tuning on the global model, resulting in a local model $\{\omega_k, e_k\}$. At this point, we perform parameter averaging between the global model and the local model. Here, $\omega_{k,p} = 0.5 \times (\omega^* + \omega_k)$ is directly used as the CRNN component of the final personalized model. As thorough global model training has already taken place, the locally fine-tuned model will not deviate significantly from the global model. Consequently, the averaged parameters $\omega^{k,p}$ will possess both global and local model's feature extraction capabilities.

Next, to enable the personalized model to emphasize local personalized features on top of the existing global model, we keep the parameters of the CRNN component $\omega^{k,p}$ fixed and proceed to fine-tune the attention layer:
\begin{linenomath}
\begin{align}
\label{ecat}
e_{k,p} = \underset{e}{\arg \min } F_k (\omega_{k,p},e)+\lambda*\|e-e^*\|
\end{align}
\end{linenomath}
where the second term represents a regularization, corresponding to $\mathcal{L}_{reg}$ in Figure~\ref{framework}, aiming to prevent the attention layer from deviating significantly from the global model.

   \begin{figure}[t]
 \centering
 \includegraphics[width=.92\linewidth]{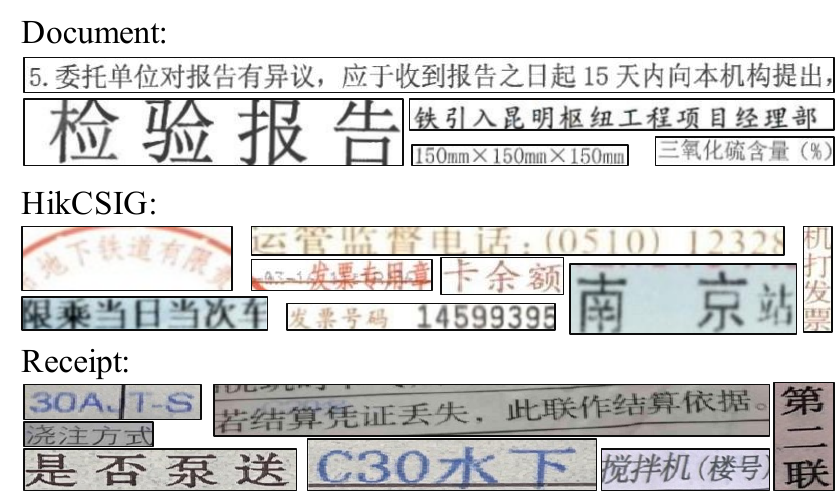}
 \caption{
Data samples. Different datasets exhibit significant stylistic differences.
 }
 \label{dataexample}
 \end{figure}

\begin{table}[]
\centering
\caption{The performance (\%) of three-client setting.  `Personalized performance' refers to the performance of each client's personalized model on their local data, while `Cross-client generalization' denotes the performance of each client's model on data from other clients.}
\resizebox{1\linewidth}{!}{
\begin{tabular}{c|cccc|ccccccc}
\Xhline{1pt} 
              \rowcolor{gray}   & \multicolumn{4}{c|}{Personalization performance} & \multicolumn{7}{c}{Cross client generalization}                                       \\ \hline
               \rowcolor{gray}  & D     & H     & \multicolumn{1}{c|}{R}    & Avg  & D$\to$H & D$\to$R & H$\to$D & H$\to$R & R$\to$D & \multicolumn{1}{c|}{D$\to$H} & Avg  \\ \hline
local           & 54.6  & 53.4  & \multicolumn{1}{c|}{59.1} & 55.7 & 5.5     & 0.4     & 16.6    & 18.8    & 8.3     & \multicolumn{1}{c|}{3.8}     & 8.9  \\
local\_p  & 55.3  & 53.4  & \multicolumn{1}{c|}{58.9} & 55.9 & 6.3     & 2.9     & 15.0    & 17.5    & 9.6     & \multicolumn{1}{c|}{2.8}     & 9.0  \\
local\_a & 55.1  & 53.7  & \multicolumn{1}{c|}{59.0} & 55.9 & 6.1     & 2.9     & 17.0    & 20.5    & 8.0     & \multicolumn{1}{c|}{3.2}     & 9.6  \\ \hline
FedAvg          & 63.0  & 55.2  & \multicolumn{1}{c|}{65.2} & 61.1 & 55.2    & 65.2    & 63.0    &  \textbf{65.2}    & 63.0    & \multicolumn{1}{c|}{ \textbf{55.2}}    & 61.1 \\
FedProx         & 62.4  & 53.5  & \multicolumn{1}{c|}{64.1} & 60.0 & 53.5    & 64.1    & 62.4    & 64.1    & 62.4    & \multicolumn{1}{c|}{53.5}    & 60.0 \\ \hline
FedAvg-ft       & 63.3  & 56.2  & \multicolumn{1}{c|}{65.3} & 61.6 & 50.0    & 60.5    & 53.8    & 52.2    & 50.0    & \multicolumn{1}{c|}{39.3}    & 51.0 \\
FedProx-ft      & 62.8  & 54.9  & \multicolumn{1}{c|}{66.2} & 61.3 & 49.0    & 59.9    & 55.9    & 55.3    & 53.4    & \multicolumn{1}{c|}{39.4}    & 52.2 \\
FedBN           & 61.2  & 56.0  & \multicolumn{1}{c|}{65.7} & 61.0 & 47.6    & 59.0    & 52.2    & 52.2    & 49.8    & \multicolumn{1}{c|}{39.6}    & 50.1 \\
FedALA          & 60.6  & 55.6  & \multicolumn{1}{c|}{64.0} & 60.1 & 41.7    & 49.6    & 37.6    & 39.8    & 36.1    & \multicolumn{1}{c|}{27.5}    & 38.7 \\
FedRep          & 55.1  & 52.7  & \multicolumn{1}{c|}{59.6} & 55.8 & 7.1     & 13.3    & 15.8    & 18.8    & 10.7    & \multicolumn{1}{c|}{5.9}     & 11.9 \\
Ditto           & 61.9  & 55.9  & \multicolumn{1}{c|}{64.2} & 60.7 & 48.1    & 57.6    & 49.3    & 51.5    & 45.7    & \multicolumn{1}{c|}{35.6}    & 48.0 \\
pFedCR  &  \textbf{71.0}  &  \textbf{58.6}  & \multicolumn{1}{c|}{ \textbf{70.1}} &  \textbf{66.6 }&  \textbf{55.3 }   &  \textbf{65.6}    &  \textbf{67.0 }   & {63.6}    &  \textbf{64.5}    & \multicolumn{1}{c|}{ 52.4}    & \textbf{61.4} \\ \Xhline{1pt} 
\end{tabular}}
\label{table1}
\end{table}

     \begin{table}[t]
\centering
\caption{The personalized performance (\%)of the 9-client setting.}
\resizebox{1.\linewidth}{!}{
\begin{tabular}{c|ccccccccc|c}
\Xhline{1pt}\rowcolor{gray}
                                                                    & c1       & c2   & c3   & c4   & c5   & c6   & c7   & c8   & c9   & Avg           \\ \Xhline{1pt}
 local           & 50.5          & 51.8      & 50.8      & 50.5      & 50.4      & 49.5      & 56.9      & 56.4      & 56.4      & 52.6          \\
local\_p  & 53.0          & 51.9      & 53.8      & 50.1      & 50.0      & 49.3      & 57.2      & 57.0      & 56.4      & 53.2          \\
 local\_a & 50.4          & 52.7      & 51.6      & 51.8      & 51.2      & 49.5      & 57.6      & 56.8      & 56.9      & 53.2          \\ \hline
 FedAvg          & 57.8          & 58.9      & 58.6      & 53.2      & 52.8      & 51.9      & 62.9      & 61.8      & 61.4      & 57.7          \\
 FedProx         & 57.0          & 58.3      & 57.5      & 51.2      & 51.1      & 50.0      & 58.8      & 58.2      & 57.8      & 55.5          \\ \hline
 FedAvg-ft       & 58.8          & 60.2      & 59.0      & 54.2      & 53.9      & 53.1      & 64.9      & 63.7      & 63.3      & 59.0          \\
 FedProx-ft      & 57.7          & 59.0      & 58.5      & 52.5      & 52.5      & 51.5      & 62.9      & 62.5      & 61.7      & 57.6          \\
 FedBN           & 57.9          & 59.0      & 58.0      & 53.7      & 53.2      & 52.9      & 63.6      & 63.1      & 62.2      & 58.2          \\
 FedALA          & 59.0          & 59.9      & 60.0      & 53.9      & 53.3      & 53.2      & 63.5      & 63.0      & 62.7      & 58.7          \\
FedRep          & 51.2          & 52.5      & 51.9      & 49.6      & 49.5      & 48.9      & 58.3      & 58.1      & 57.5      & 53.1          \\
 Ditto           & 57.5          & 58.5      & 57.9      & 53.7      & 53.5      & 52.7      & 63.6      & 62.6      & 62.0      & 58.0          \\
 pFedCR  & \textbf{66.2} & \textbf{66.3} & \textbf{66.1} & \textbf{55.2} & \textbf{55.0} & \textbf{54.3} & \textbf{66.9} & \textbf{65.8} & \textbf{65.8} & \textbf{62.4 } \\ \Xhline{1pt}
\end{tabular}}
\label{9client}
\end{table}

\section{Experiments}

In this section, we conduct a series of experiments to answer the following questions: 1) How much gain can pFL bring to Chinese text recognition? 2) How does our proposed method compare with other pFL methods? 3) What are the key factors that facilitate the performance in our proposed approach?

\subsection{Datasets}
In this paper, we conduct experiments on three datasets, including two collected datasets (\textit{i.e.}, Document and Receipt) and one from CSIG 2022 Competition\footnote{https://davar-lab.github.io/competition/CSIG2022-invoice.html} named HikCSIG. The text images of the Document are easier to recognize since most of them are scanned. On the contrary, samples of Receipts are all captured by mobile phone, resulting in most of the samples being accompanied by perspective and blurring. The dataset HikCSIG contains six types of invoices including air tickets,  general quota and taxi invoices, passenger transport invoices, toll invoices, and train tickets. More details of the adopted three datasets are in Supplementary materials, and some samples are shown in Figure \ref{dataexample}.


For federated learning, we adopt two data partitioning methods. One is to treat each dataset as an individual client, where style and character category differences exist among these clients. The other considers the fact that in real-world scenarios, institutions with similar data distributions may exist. Thus, we randomly divide the train and test sets of each dataset into three groups, forming a total of nine clients.

\subsection{Implementation Details}
We implement our method in PyTorch~\cite{paszke2019pytorch} and conduct all experiments on a single NVIDIA GeForce RTX 3090 GPU. We use the Adadelta optimizer with an initial learning rate of 1, and the learning rate decay follows the cosine schedule in different federated learning rounds. By default, the global communication round $T$ is set to 35, and the local batch size during training is set to 128. In each communication round, the local training epoch and the server-side fine-tune epoch are set to 1. 

\subsection{Comparison}
We compare pFedCR with different types of methods:
1)  \textbf{local}, \textbf{local\_a} and \textbf{local\_p}, where \textbf{local} means standalone training without federated learning, and each client trains its own model to convergence; \textbf{local\_a} means train CRNN with ECA layers; \textbf{local\_p} means pre-training with the same virtual data used on the server. 
2) Standard federated learning methods: \textbf{FedAvg}~\cite{mcmahan2017communication} and \textbf{FedProx}~\cite{conf/mlsys/LiSZSTS20}, which aim to learn a unified global model for all clients. FedProx adds an additional regularization loss to the local training based on FedAvg to prevent local models from deviating too far from the global model. 
3) Existing pFL methods: \textbf{FedAvg-ft} and \textbf{FedProx-ft} represent fine-tuning the client models for one extra epoch based on the global model to obtain personalized models. \textbf{FedBN}~\cite{li2021fedbn} allows each client to retain local personalized BN layers based on FedAvg.  \textbf{FedALA}~\cite{zhang2022fedala} studies how to personalize the selection of global model parameters as initialization before local training. \textbf{FedRep}~\cite{collins2021exploiting} divides the model into basic and personalized layers, with only the parameters of the basic layer uploaded to the server for parameter averaging. \textbf{Ditto}~\cite{li2021ditto} conducts additional personalized training locally, and uses global model parameters for regularization.

\subsection{Main Results} 
Table~\ref{table1} shows the results of the three-client setting. In order to provide a comprehensive comparison of different methods, we present both personalized performance and cross-client generalization performance. Personalized performance refers to the performance of the personalized model on the local test set of each client after participating in federated learning. Cross-client generalization performance refers to the performance of a model trained by one client on the test set of other clients. $D\to H$ represents the performance of the \textit{Document} client model in the \textit{HikCSIG} client after the end of federated learning. From Table~\ref{table1}, we can see that:

\begin{itemize}[noitemsep, leftmargin=*]
\item Simply adding the attention layers to CRNN has a weak effect on improving model performance. Moreover, using virtual data on the server for model pre-training has little impact on the final performance. This indicates that operations performed within clients are difficult to break through the client performance bottleneck.

\item Through federated learning, both standard and personalized methods can significantly improve model performance, including personalized performance and cross-client generalization performance. This reflects the necessity of collaborative learning in Chinese text recognition. As we will reveal in subsequent experiments, federated learning can significantly improve the accuracy of client models on few-shot characters.

\item Although standard federated learning has made significant improvements compared to local training, ignoring the personalized requirements of different clients results in significant room for improvement in local client data.

\item Existing pFL methods do not significantly outperform standard federated learning methods in Chinese text recognition. Moreover, the simplest method, FedAvg-ft, achieves the best results in these personalized methods.

\item Overall, the personalized performance of pFedCR is better than all other comparative methods. In addition, the personalized performance is 7\% higher than that of the second place. Furthermore, in terms of cross-client generalization performance, pFedCR outperforms all compared methods.

\end{itemize}

  \begin{figure}[t]
 \centering
 \includegraphics[width=1\linewidth]{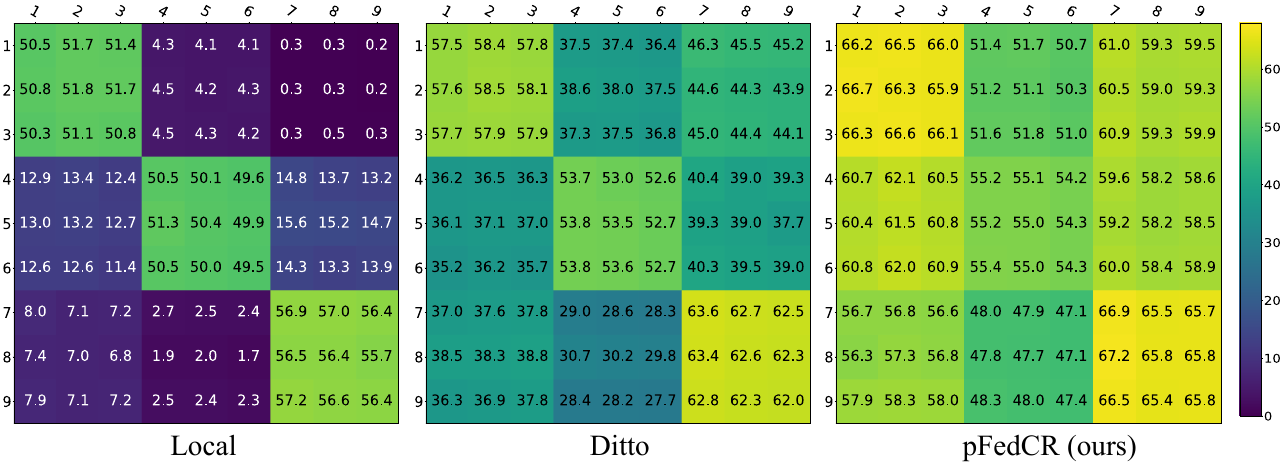}
 \caption{
Cross-client performance in the nine-client setting.
 }
 \label{gener}
 \end{figure}

\begin{table}[]
\centering
\caption{Ablation study. The impact of each component operation in pFedCR on the final results.}
\resizebox{.95\columnwidth}{!}{
\begin{tabular}{cccc|ccc|c}
\Xhline{1pt}\rowcolor{gray}
\begin{tabular}[c]{@{}c@{}}Virtual \\ data\end{tabular} & \begin{tabular}[c]{@{}c@{}}ECA \\ layer\end{tabular} & Freeze     & Stage 2    & client1 & client2 & client3 & Avg           \\ \Xhline{1pt}
\XSolidBrush & \XSolidBrush  & \XSolidBrush &  \XSolidBrush & 63.0    & 55.2    & 65.2    & 61.1          \\ \hline
\Checkmark                                              &   \XSolidBrush    & \XSolidBrush  &   \XSolidBrush  & 69.5    & 57.7    & 67.6    & 64.9          \\ \hline
\Checkmark                                              & \Checkmark                                           &   \XSolidBrush  &  \XSolidBrush& 69.5    & 58.2    & 67.3    & 65.0          \\ \hline
\Checkmark                                              & \Checkmark                                           & \Checkmark &\XSolidBrush  & 71.6    & 58.7    & 68.4    & 66.2          \\ \hline
\Checkmark                                              & \Checkmark                                           & \Checkmark & \Checkmark & 71.0    & 58.6    & 70.1    & \textbf{66.6} \\\Xhline{1pt}
\end{tabular}}
\label{aba}
\end{table}

Table~\ref{9client} shows that pFedCR still maintains the best personalized performance in the setting of nine clients. Each client in collaborative learning has greatly improved performance. Compared with not conducting federated learning, pFedCR can improve the average performance by 18.63\%. Compared with the existing best personalized FL methods, pFedCR improves the average performance by 5.7\%.

In Figure~\ref{gener}, the cross-client generalization performance of different models in the nine-client setting is shown. We present the performance of personalized models obtained by four different methods across all clients. The darker the color, the worse the performance. It can be seen that \textbf{1)} when not conducting federated learning, the model can only perform well in clients with the same data distribution, while it completely fails in other distributions. On the contrary, when conducting pFL, the personalized model of each client not only improves its local performance but also significantly improves its performance in handling other distribution data. \textbf{2)} Among several pFL methods, pFedCR not only achieves the best personalized performance but also achieves the best performance in handling cross-client data.

\subsection{Ablation Study}
We evaluate the effects of different operations. Specifically, we decompose pFedCR into four components: adding virtual data, adding ECA layers, freezing the last embedding layer, and incorporating the second-stage client fine-tuning. Table~\ref{aba} displays the results of different combinations of these four components. It is evident that adding virtual data on the server has an intuitively significant impact on the final performance. Moreover, the other operations also collectively improve the performance of pFedCR.

 \begin{figure}[t]
 \centering
 \includegraphics[width=0.8\linewidth]{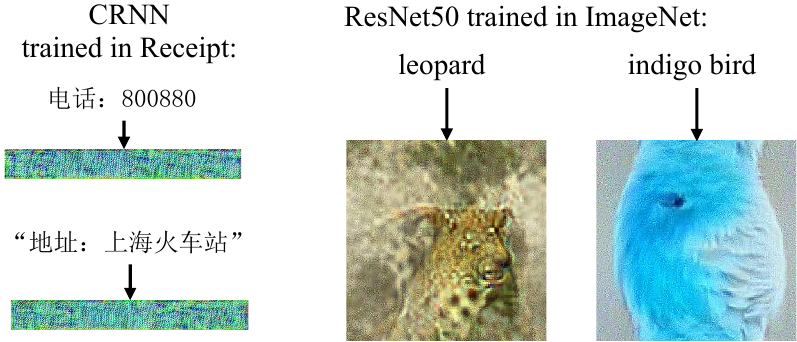}
 \caption{
Reconstruct Images from Model Parameters using DeepInversion. Even with labels provided, it is hard to reconstruct strings from CRNN.
 }
 \label{attack}
 \end{figure} 
 
\subsection{Discussion on Privacy}
We only share model parameters between the server and clients. Then one question is, can character information, such as telephone numbers, be inferred from model parameters? We note that in federated image classification, some studies~\cite{yin2020dreaming,yin2021see} have successfully reconstructed class-specific images from model parameters. However, in our work, the privacy concern resides within characters. The reconstruction of characters from model parameters is nearly impossible. To validate this point, we employ DeepInversion~\cite{yin2020dreaming} to separately attempt image reconstruction from both a pre-trained ResNet-50 on ImageNet and the CRNN trained in our work. Figure~\ref{attack} illustrates that even when feeding labels to DeepInversion, retrieving textual information from CRNN is challenging. Conversely, obtaining class-related information from ResNet-50 is straightforward.

\subsection{Analysis}

\begin{table}[]
\centering
\caption{The performance (\%) on characters from different frequency intervals.}
\resizebox{.92\columnwidth}{!}{
\begin{tabular}{cc|cccccc}
\Xhline{1pt}\rowcolor{gray}
                                              &           & \multicolumn{6}{c}{Different frequencies}       \\ \cline{3-8} 
                                              &           & 0-0  & 1-10 & 11-20 & 21-30 & 200-400 & 400-800 \\ \hline
\multicolumn{1}{c|}{\multirow{3}{*}{Document}} & Local     & 0.0  & 14.6 & 30.2  & 32.2  & 65.4    & 71.1    \\
\multicolumn{1}{c|}{}                         & FedAvg-ft & 31.6 & 29.2 & 52.2  & 64.8  &  \textbf{83.6}    &  \textbf{87.5}     \\
\multicolumn{1}{c|}{}                         & Ours      &  \textbf{61.6}  &  \textbf{50.9}  & 64.7  & \textbf{65.8}   & 80.9    & 83.2    \\ \hline
\multicolumn{1}{c|}{\multirow{3}{*}{HikCSIG}} & Local     & 0.0  & 0.7  & 19.0  & 30.3  & 58.2    & 67.1    \\
\multicolumn{1}{c|}{}                         & FedAvg-ft & 11.1 & 13.1 & 23.2  & 34.0  &  \textbf{65.2}    &  \textbf{76.4}     \\
\multicolumn{1}{c|}{}                         & Ours      &  \textbf{27.4}  &  \textbf{30.9}  &  \textbf{32.5}   &  \textbf{38.9}   & 63.8    & 74.9    \\ \hline
\multicolumn{1}{c|}{\multirow{3}{*}{Receipt}} & local     & 0.0  & 0.4  & 30.4  & 43.7  & 52.9    & 67.2    \\
\multicolumn{1}{c|}{}                         & FedAvg-ft & 26.9 & 18.0 & 64.9  & 71.8  &  \textbf{71.5}     & 85.6    \\
\multicolumn{1}{c|}{}                         & Ours      &  \textbf{47.5}  &  \textbf{34.1}  &  \textbf{64.1}   &  \textbf{74.7}   & 68.9    &  \textbf{88.0}     \\\Xhline{1pt}
\end{tabular}}
\label{fewshot}
\end{table}

\textbf{Performance improvement in few shot characters.}
To demonstrate the performance variation in few-shot characters, we count the occurrence frequency of each character in the training set of each client under the three-client setting, and divide these characters into several frequency intervals. In Table~\ref{fewshot}, the interval `0-0' represents characters that are not included in the local trainset. In addition to presenting four intervals of low-frequency characters, we also present two intervals of high-frequency characters, namely `200-400' and `400-800'. It shows that: 1) pFL enhances few-shot character performance compared to standalone training, with notable gains in high-frequency characters. 2) pFedCR outperforms FedAvg-ft significantly in few-shot character performance, with minimal high-frequency character loss.

 \begin{figure}[t]
 \centering
 \includegraphics[width=0.7\linewidth]{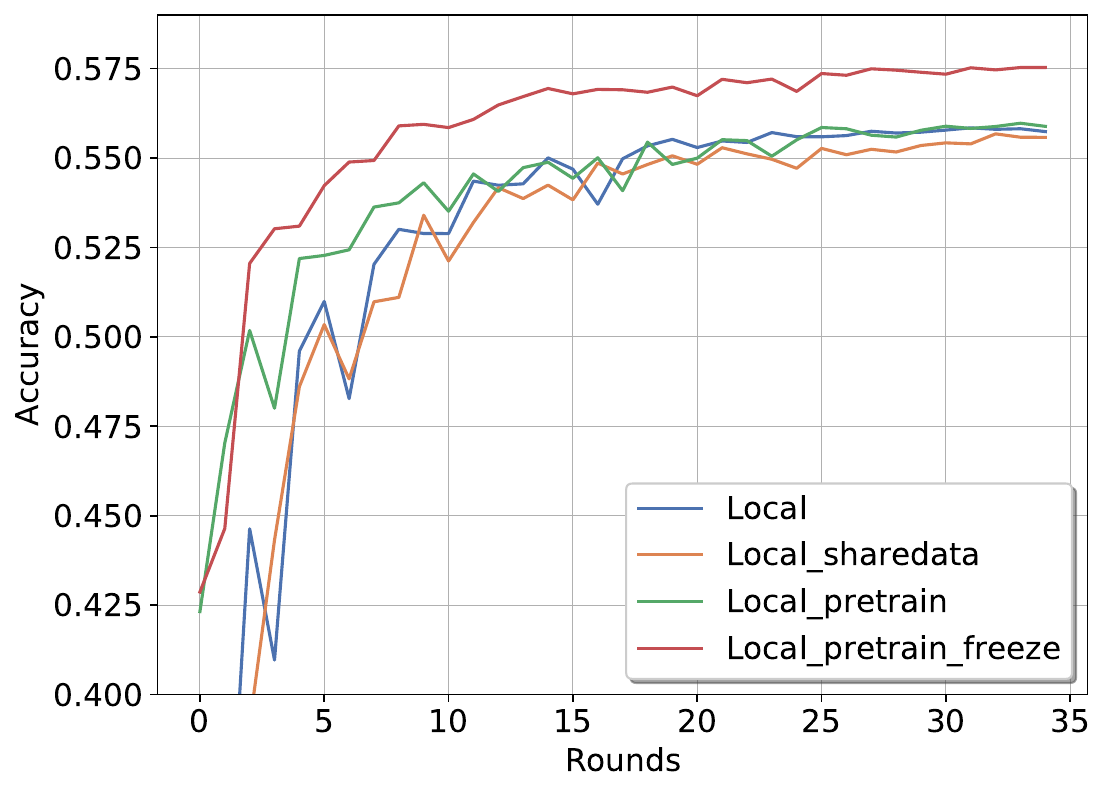}
 \caption{
The impact of virtual data on standalone training.
 }
 \label{localtrain}
 \end{figure} 
 
 \textbf{Why is it difficult for standalone training to benefit from pretraining?}
To understand the slight improvement of pre-training on standalone training in Tables~\ref{table1} and \ref{gener}, we conduct a set of comparative experiments. As shown in Figure~\ref{localtrain}, `local\_sharedata' shares server-generated virtual data among all clients for independent training; `local\_pretrain' initializes all client models with a pre-trained model from virtual data; `local\_pretrain\_freeze' initializes the client model with pre-trained weights, freezing the last layer during training. The results show that although the model's performance improves in the early stage after adding pre-training, there is no significant improvement after the model converges. Remarkably, freezing the last layer leads to notable local model performance changes, highlighting the efficacy of balanced embedding layer training.

 

\subsection{Visualization}
\begin{figure}[t]
 \centering
 \includegraphics[width=0.95\linewidth]{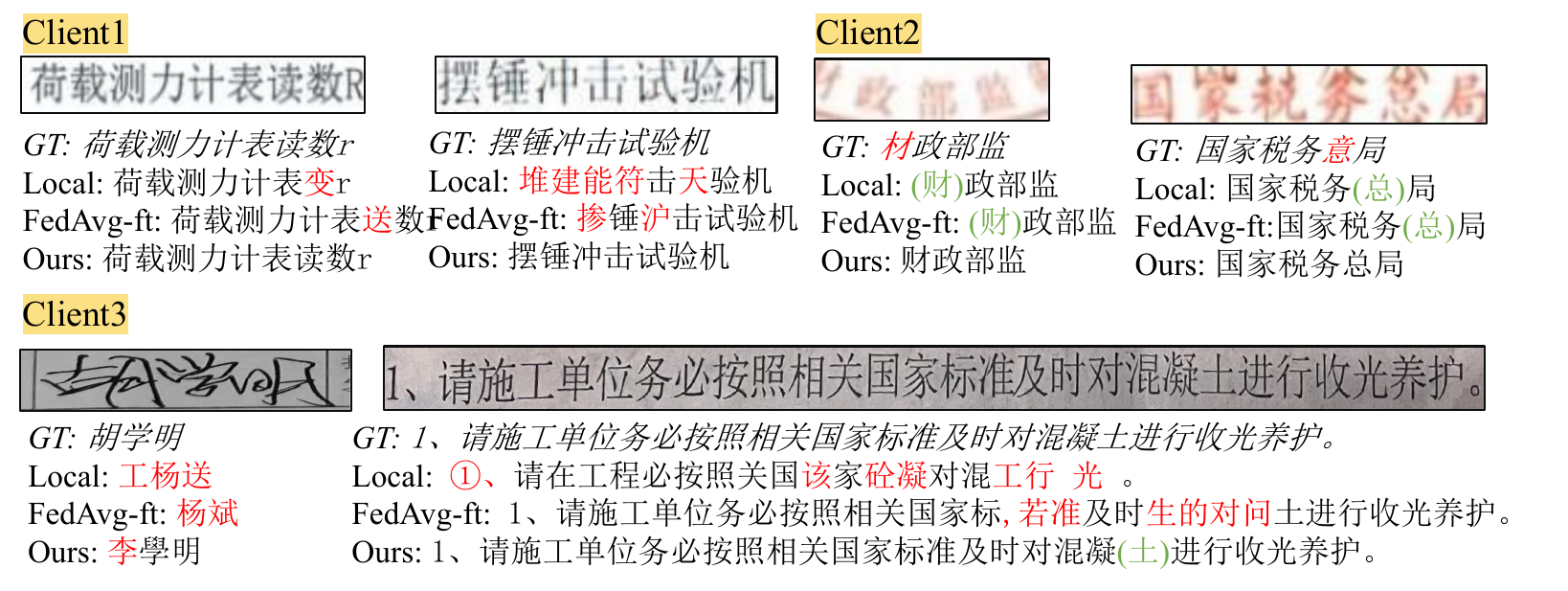}
 \caption{
Visualization of the personalized performance.
 }
 \label{gen}
 \end{figure}

\begin{figure}[t]
 \centering
 \includegraphics[width=0.95\linewidth]{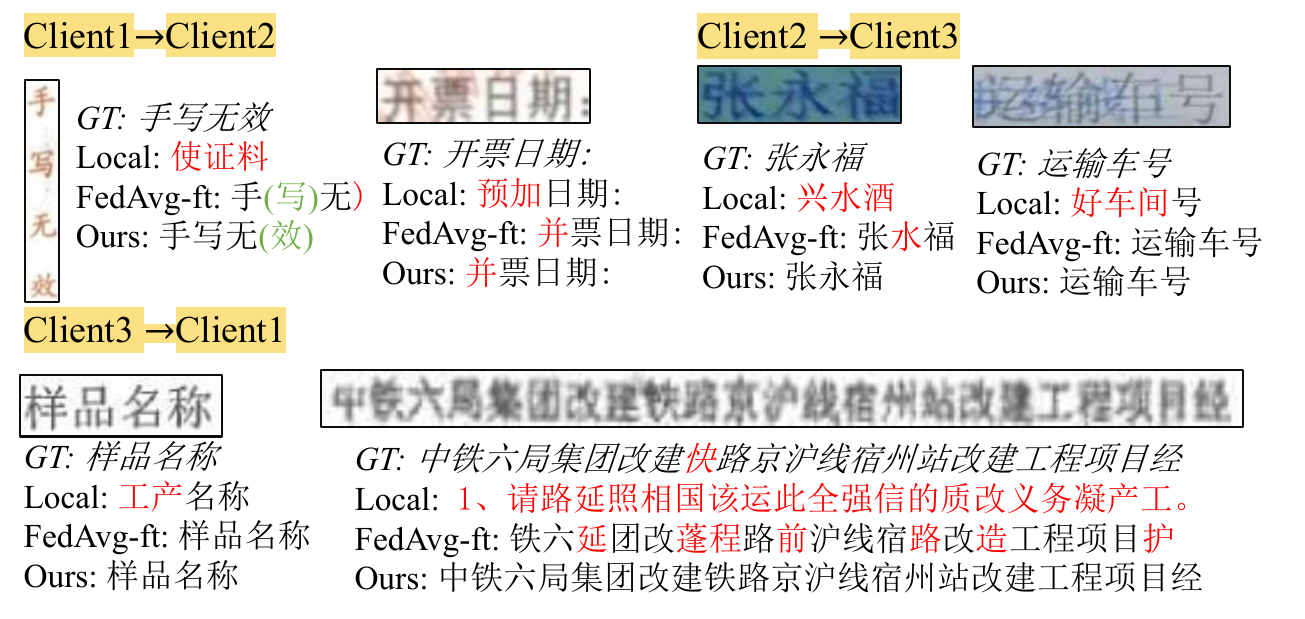}
 \caption{
Visualization of the cross-client generalization.
 }
 \label{gen2}
 \end{figure}

We visualize the final prediction results of the personalized models under the three-client setting. Figure~\ref{gen} shows the prediction results of each client's personalized model, and Figure~\ref{gen2} shows the generalization performance of the personalized models on cross-client data. Red indicates incorrect characters, while green indicates missing characters. The prediction results show that: 1) pFL can not only improve local performance but also improve generalization across clients, demonstrating the potential application of collaborative learning in Chinese text recognition with different distribution data. 2) Our proposed method has significant performance improvements compared to standalone training and other pFL methods. \textbf{It can correct the noise labels in the dataset, such as client2 in Figure~\ref{gen} and client3$\to$client1 in Figure~\ref{gen2}}. This indicates that in federated learning, not only the feature extraction ability of the CNN part is enhanced, but also the relationship extraction ability of the RNN part is further improved.

\section{Conclusions}
In this paper, we explore collaborative Chinese text recognition and propose a pFL method, pFedCR, which consists of two stages: global model training and personalization. To address the issue of character imbalance, we introduce a virtual data fine-tuning step on the server to improve the standard parameter averaging. 
To better adapt the personalized model to different feature distributions, we introduce an attention mechanism into the model and perform regularization fine-tuning during Stage 2. Extensive experimental results demonstrate that pFL can significantly improve the performance of client models, including both personalized and cross-client generalization performance. Moreover, pFedCR outperforms all the comparative methods.
%

\bibliography{aaai24}

\end{document}